\documentclass[10pt, conference, compsocconf]{IEEEtran}
\usepackage{cite}

\ifCLASSINFOpdf
\else
\fi

\usepackage{cite}
\usepackage{url}
\usepackage{subfigure}
\usepackage{epsfig}
\usepackage{amsmath}
\usepackage{verbatim}
\usepackage{color}
\usepackage{listings}
\usepackage{multirow}
\usepackage[usenames,dvipsnames]{xcolor}
\usepackage{booktabs}
\usepackage{float}

\newcommand{\argmin}{\mathop{\mathrm{argmin}}}

\newcommand{\eat}[1]{}

\begin{document}
\title{A Hypergraph-Partitioned Vertex Programming Approach \\
for Large-scale Consensus Optimization}


\author{
    \IEEEauthorblockN{Hui Miao\IEEEauthorrefmark{1}, Xiangyang Liu\IEEEauthorrefmark{2}, Bert Huang\IEEEauthorrefmark{1}, Lise Getoor\IEEEauthorrefmark{1}}
    \IEEEauthorblockA{\IEEEauthorrefmark{1}Dept. of Computer Science, \IEEEauthorrefmark{2}Dept. of Electrical \& Computer Engineering\\
    University of Maryland, College Park, USA\\
    \{hui, bert, getoor\}@cs.umd.edu\IEEEauthorrefmark{1}, xyliu@umd.edu\IEEEauthorrefmark{2}}

}

\newcommand{\secref}[1]{Section~\ref{#1}}
\newcommand{\appref}[1]{Appendix~\ref{#1}}
\newcommand{\figref}[1]{Fig.~\ref{#1}}
\newcommand{\tabref}[1]{Table~\ref{#1}}
\newcommand{\eqnref}[1]{equation~\ref{#1}}

\newcommand{\commentout}[1]{%
}

\newcommand{\from}[3]{{\textbf{FROM #1 TO #2: #3}}}
\newcommand{\todo}[1]{\textbf{TODO: #1}}
\newcommand{\noshow}{{}}

\newcommand{\ACO}{{\sc aco}} 
\newcommand{\ACOH}{{\sc aco-Hyper}} 
\newcommand{\ACOG}{{\sc aco-Greedy}} 
\newcommand{\ACOR}{{\sc aco-Rand}} 
\newcommand{\hyper}{{\sc Hyper}}
\newcommand{\random}{{\sc Random}}
\newcommand{\greedy}{{\sc Greedy}}

\maketitle

\begin{abstract}
In modern data science problems, techniques for extracting value from big data require
performing large-scale optimization over 
heterogenous, irregularly structured data.  Much of this data is best
represented as multi-relational graphs, making vertex programming
abstractions such as those of Pregel and GraphLab ideal fits for
modern large-scale data analysis.  In this paper, we describe a
vertex-programming implementation of a popular consensus optimization
technique known as the \emph{alternating direction of multipliers}
(ADMM) \cite{Boyd:DistrubutedADMM}. 
ADMM consensus optimization allows elegant solution of
complex objectives such as inference in rich probabilistic models.
We also introduce a novel hypergraph partitioning technique
that improves over state-of-the-art partitioning techniques for
vertex programming and significantly reduces the communication cost
by reducing the number of replicated nodes up to an order of magnitude.
We implemented our algorithm in GraphLab 
and measure scaling performance on a variety of realistic bipartite graph distributions
and a large synthetic voter-opinion analysis application.  
In our experiments, we are able to achieve a 50\% improvement in
runtime over the current state-of-the-art GraphLab
partitioning scheme.
\end{abstract}


%
\IEEEpeerreviewmaketitle

\section{Introduction}


Large-scale data often contains noise, statistical dependencies, and
complex structure. 
To extract value from such data, we need both flexible, expressive models 
and scalable algorithms to perform reasoning over these models.
In this paper, we show how a general class of distributed optimization techniques
can be implemented efficiently on 
graph-parallel abstractions frameworks.

Consensus optimization using \emph{alternating method of multipliers} (ADMM), 
is a recently popularized general method for distributed solution of
large-scale complex optimization problems. 
The optimization problem is decomposed into simple subproblems to be
solved in parallel.
The combination of decomposability of dual ascent and the
fast convergence of method of multipliers makes it suitable for many
problems including distributed signal
processing \cite{Ernie:ADMM-Bregman},
inference in graphical
models \cite{Tziritas:MRFDD,bach:nips12}, and 
popular machine learning algorithms \cite{Georgios:ADMM-SVM}.

\emph{Vertex programming} is an efficient graph-parallel abstraction
for distributed graph computation. Pregel \cite{Czajkowski:Pregel},
Giraph \cite{giraph:tool} and GraphLab \cite{gonzalez2012powergraph}
are recently proposed frameworks for parallelizing graph-intensive
computation. These frameworks adopt a vertex-centric model to define
independent programs on each vertex. Experiments show it outperforms
the MapReduce abstraction by one to two orders of magnitude in machine learning and
data mining algorithms \cite{YuchengLow:DistributedGraphlab}.

In this paper, 
we investigate the bipartite topology of general ADMM-based consensus
optimization and develop its vertex-programming formulation.  We also
propose a novel partitioning scheme that utilizes the 
characteristics of computation graph based on a hypergraph
interpretation of the bipartite data graph. Our partitioning can
reduce the number of replicated vertices by an order of magnitude over the 
current state-of-the-art partitioning scheme, reducing communication cost
accordingly.   

Our contributions include the following:
\begin{itemize}
\item We develop a scalable, parallel algorithm for
ADMM-based consensus optimization using the vertex programming abstraction. The underlying computation
graph takes on a bipartite structure. 
\item We propose a partitioning method that treats the bipartite graph
as a hypergraph and performs a hypergraph cut to distribute
the nodes across machines. This strategy enables our
replication factor to outperform the state-of-the-art graph
partitioning scheme \cite{gonzalez2012powergraph}. The running
time of ADMM-based consensus optimization using our partition
strategy is approximately half 
of the time compared to current state-of-the-art methods.
Our partitioning strategy is of
independent interest, since it can be used for any problem
that decomposes into a bipartite computation graph, such as
factor-graph belief propagation \cite{Loeliger:sumProductFactorGraph}.
\item
We implement our algorithm in GraphLab and evaluate our partitioning scheme on a social network
analysis problem, demonstrating that we can perform large-scale
probabilistic inference on a modest number of machines, even when the
graph has the heavy degree skew often found in real networks.
\end{itemize}

\commentout{
Large-scale data often contains noise, statistical dependencies, and
complex structure. To reason about such data, one needs flexible and
expressive probabilistic models and scalable algorithms to perform
inference over these models.
\emph{Consensus optimization} is a recently popularized general method for distributed solution of complex large-scale probabilistic inference. Using the \emph{alternating method of multipliers} (ADMM), the complex consensus optimization can be decomposed into simple subproblems to be solved in parallel. 
\emph{Vertex programming} is a powerful framework for distributed computing for graph-structured problems.
In this paper, we develop the first vertex-programming formulation of consensus optimization that uses a novel partitioning scheme based on a hypergraph interpretation of the data graph. Our partitioning reduces the excess replication factor to less than 30\% of the current state-of-the-art partitioning scheme, reducing communication cost accordingly. 

Our contributions are as follows:
\begin{enumerate}
	\item We develop a scalable, parallelized algorithm for ADMM-based consensus optimization. The underlying computation graph takes on a bipartite structure.  
	\item We propose a partitioning method for bipartite graphs that treats the graph as a hypergraph and performs a hypergraph cut to distribute the nodes across machines. This strategy causes our replication factor to outperform the state-of-the-art graph partitioning scheme \cite{gonzalez2012powergraph}. The running time of ADMM-based consensus optimization using our partition strategy is reduced to approximately half of the time compared to using vertex partition. Our partitioning strategy is of independent interest, since it can be used for any problem that decomposes into a bipartite computation graph, such as factor-graph belief propagation 
	\cite{Loeliger:sumProductFactorGraph}.
\end{enumerate}

We evaluate our algorithm and partitioning scheme on a social network analysis problem, demonstrating that we can perform large-scale probabilistic inference on a modest number of machines, even when the graph has the heavy degree skew often found in real networks.

}

\section{Motivation}
\label{sec:motivation}

Before describing our proposed algorithm (\secref{sec:algorithm}), we begin
with a simple illustrative example of the kind of problems that can be
solved using ADMM.  We consider the task of social network
analysis on a network of individuals connected by various social
relationships. The goal of the analysis is to predict the
voting preferences of various individuals in the social network, using
the relationships and some observed voting preferences of other individuals
in the network.
The problem can be cast as a probabilistic inference problem, and
the approach that we take here is to define the model 
using \emph{probabilistic soft logic} (PSL) \cite{Lise:shortPSL}.
PSL is a general-purpose language for describing
large-scale probabilistic models over continuous-valued random variables
using weighted logical rules. 

A PSL program consists of a set of logical rules with conjunctive
bodies and disjunctive heads (negations allowed). 
Rules are labeled with non-negative weights. The
following program, based on an example from Bach et al.~\cite{bach:nips12}, 
encodes a simple model to predict voter behavior using
information about a voter (voter registration) and their
social network described by two types of links indicating
$\textsc{Friend}$ and $\textsc{Spouse}$ relationships:
\begin{align}
&\small\textup{1.1:} &\small\textsc{RegisteredAs}\small\textit{(A, P)} &\rightarrow \small\textsc{Votes}\small\textit{(A,P)},\nonumber\\
&\small\textup{0.5:} &\small\textsc{Friend}\small\textit{(B, A)} \wedge \small\textsc{Votes}\small\textit{(A,P)} &\rightarrow \small\textsc{Votes}\small\textit{(B,P)},\nonumber\\
&\small\textup{1.8:} &\small\textsc{Spouse}\small\textit{(B, A)} \wedge \small\textsc{Votes}\small\textit{(A,P)} &\rightarrow \small\textsc{Votes}\small\textit{(B,P)}.\nonumber
\end{align}
Consider any
constants for persons, $a$ and $b$ and party $p$ instantiating logical terms
$A$, $B$, and $P$ respectively. 
The first rule encodes the correlation between voter
registration and party preferences, which tend to be aligned but are not
always. 
The next rule states that if $a$ is a friend of
$b$ and votes for party $p$, there is a chance that $b$ votes for party $p$ as
well, whereas the second makes the same statement for spouses. The rule weights
indicate that spouses are more likely to vote for the same party than
friends.
The resulting probabilistic model will combine all of these influences
and include the implied structured dependencies. 
We can also consider more rules and relationship types, leading to a full program in Figure \ref{fig:pslprogram}. PSL can also include constraints on logical atoms, such as mutual exclusivity of voting preferences \textsc{Votes}.

\begin{figure}[t]
\vspace{-20pt}
\begin{align*}
&\small\textup{0.5:} &&\small\textsc{RegisteredAs}\small\textit{(A, P)} \hspace{-0.1in} &\rightarrow \small\textsc{Votes}\small\textit{(A,P)},\nonumber\\
&\small\textup{0.3:} &\small\textsc{Votes}\small\textit{(A, P)} &\wedge \small\textsc{KnowsWell}\small\textit{(B, A)} \hspace{-0.1in} &\rightarrow \small\textsc{Votes}\small\textit{(B,P)},\nonumber\\
&\small\textup{0.1:} &\small\textsc{Votes}\small\textit{(A, P)} &\wedge \small\textsc{Knows}\small\textit{(B, A)} \hspace{-0.1in} &\rightarrow \small\textsc{Votes}\small\textit{(B,P)},\nonumber\\
&\small\textup{0.05:} &\small\textsc{Votes}\small\textit{(A, P)} &\wedge \small\textsc{Boss}\small\textit{(B, A)} \hspace{-0.1in} &\rightarrow \small\textsc{Votes}\small\textit{(B,P)},\nonumber\\
&\small\textup{0.1:} &\small\textsc{Votes}\small\textit{(A, P)} &\wedge \small\textsc{Mentor}\small\textit{(B, A)} \hspace{-0.1in} &\rightarrow \small\textsc{Votes}\small\textit{(B,P)},\nonumber\\
&\small\textup{0.7:} &\small\textsc{Votes}\small\textit{(A, P)} &\wedge \small\textsc{OlderRelative}\small\textit{(B, A)} \hspace{-0.1in} &\rightarrow \small\textsc{Votes}\small\textit{(B,P)}.\nonumber
\end{align*}
\caption{Political social network voting program written in
  probabilistic soft logic. Additionally, the \textsc{Votes} predicate
  is constrained to have total truth value of 1.0, to
  preserve mutual exclusivity of voting preference.}\label{fig:pslprogram}
\end{figure}

The engine behind PSL compiles the logical program into a
continuous-variable representation known as a \emph{hinge-loss Markov
  random field} (HL-MRF) \cite{bach:nips12,bach:uai13}. Like many probabilistic
graphical models, inference in HL-MRFs can be distributed and solved
using consensus optimization. In HL-MRFs, inference of the
most-probable explanation (MPE) is a convex optimization, since the
logical rule are converted into hinge-loss potentials and constraints
such as mutual exclusivity can be relaxed to linear equalities. Since
inference is a convex optimization, HL-MRFs are particularly
well-suited for consensus optimization.
We defer to previous papers for the mathematical formalisms of
HL-MRFs, and here mainly discuss the general implementation of
ADMM-based consensus optimization, which has many applications beyond PSL.

\commentout{
In this section, we describe an example use case of our algorithm, following the setup described by Bach et al.~\cite{bach:nips12}. 
We consider the task of social network analysis on a network of individuals connected by various social relationships. The goal of the analysis is to discover the unreported voting preferences of various individuals in the social network, using the relationships and some observed voting preferences in the network.

To solve this problem, we can build a probabilistic model using \emph{probabilistic soft logic} (PSL) \cite{kimmig:probprog12}, a general-purpose tool for designing and performing inference in probabilistic models defined by logical rules. 
A PSL program consists of a set of logical rules with conjunctive
bodies and disjunctive heads. Rules are labeled with non-negative weights. The
following example program encodes a simple model to predict voter behavior using
voter registration and a social network with two types of links denoting
$\textsc{Friend}$ and $\textsc{Spouse}$ relationships:
\begin{align}
0.5: \textsc{Friend}(B,A) \wedge \textsc{Votes}(A,P)&\rightarrow \textsc{Votes}(B,P)\nonumber\\
1.8: \textsc{Spouse}(B,A)  \wedge \textsc{Votes}(A,P)&\rightarrow \textsc{Votes}(B,P)\nonumber\\
1.1: \textsc{RegisteredAs}(A, P) &\rightarrow \textsc{Votes}(A,P)\label{eq:advertise}.\nonumber
\end{align}
Consider any
constants for persons, $a$ and $b$ and party $p$ instantiating logical terms
$A$, $B$, and $P$ respectively. The first rule states that if $a$ is a friend of
$b$ and votes for party $p$, there is a chance that $b$ votes for party $p$ as
well, whereas the second makes the same statement for spouses. The rule weights
indicate that spouses are more likely to vote for the same party than
friends. Finally, the last rule encodes the correlation between voter
registration and party preferences, which tend to be aligned but are not
always. The resulting probabilistic model will combine all of these influences
and include the implied structured dependencies. 
We can also consider more rules and relationship types, leading to a full program in Figure \ref{fig:pslprogram}. PSL can also include constraints on logical atoms, such as mutual exclusivity of voting preferences \textsc{Votes}.

\begin{figure*}
\begin{align*}
0.5 & :  \textsc{RegisteredAs}(A,P) \rightarrow \textsc{Votes}(A,P)\\
0.3 & :  \textsc{Votes}(A,P) \wedge \textsc{KnowsWell}(B,A) \rightarrow \textsc{Votes}(B,P)\\
0.1 & :  \textsc{Votes}(A,P) \wedge \textsc{Knows}(B,A) \rightarrow \textsc{Votes}(B,P)\\
0.05 &:  \textsc{Votes}(A,P) \wedge \textsc{Boss}(B,A) \rightarrow \textsc{Votes}(B,P)\\
0.1 & : \textsc{Votes}(A,P) \wedge \textsc{Mentor}(B,A) \rightarrow \textsc{Votes}(B,P)\\
0.7 & : \textsc{Votes}(A,P) \wedge \textsc{OlderRelative}(B,A) \rightarrow \textsc{Votes}(B,P)\\
0.8 & : \textsc{Votes}(A,P) \wedge \textsc{Idol}(B,A) \rightarrow \textsc{Votes}(B,P)\\
\end{align*}
\caption{Political social network voting program written in probabilistic soft logic. Additionally, the \textsc{Votes} predicate is constrained to only have total sum truth value of 1.0, to preserve mutual exclusivity of voting preference.}\label{fig:pslprogram}
\end{figure*}

The engine behind PSL compiles the logical program into a continuous-variable representation known as a \emph{hinge-loss Markov random field} (HL-MRF) \cite{bach:uai13}. Like many probabilistic graphical models, inference in HL-MRFs can be distributed and solved using consensus optimization. In HL-MRFs, inference of the most-probable explanation (MPE) is a convex optimization, since the logical rule are converged into hinge-loss potentials and constraints such as mutual exclusivity can be relaxed to linear equalities. Since inference is a convex optimization, HL-MRFs are particularly well-suited for consensus optimization \cite{bach:nips12,bach:uai13}.
We defer to previous papers for the mathematical formalisms of HL-MRFs, and here mainly discuss the general implementation of ADMM-based consensus optimization. 
}

\section{Preliminaries}
\subsection{ADMM-Based Consensus Optimization}
\label{sec:admm_intro_sub}

Consensus optimization simplifies solution of a global objective by decomposing a complex objective into simpler subproblems over local copies of the variables and constraining each local copy to be equal to a global \emph{consensus variable}. 
The general form of the consensus optimization is:
\begin{equation}
\begin{aligned}
& \min_{x_{1}, \ldots, x_{N}}
& & \sum_{i=1}^{N}\phi_{i}(x_{i})\\
& \text{subject to}
& &x_{i} - \vec{X_{i}} = 0,i=1,2,...,N , \\
\end{aligned}
\end{equation}
where $x_{i}$ with dimension $n_{i}$ is the local variable vector 
on which the $i$th subproblem depends and  $\phi_{i}$ is the objective function for the $i$th subproblem. For ease of notation, let $\vec{X_{i}}$ denote the global consensus variable vector that the local variable $x_{i}$ should equal.

To solve the consensus optimization, 
ADMM relaxes the global equality constraints using an augmented Lagrangian and solves the dual objective. The ADMM-based solution procedure is
\cite{Boyd:DistrubutedADMM}:
\begin{align} \label{eq:admm}
&x_{i}^{k+1} \leftarrow \argmin_{x_{i}}\left( \phi_{i}(x_{i})+ \lambda_{i}^{k}\cdot x_{i} + \frac{\rho}{2}\left\| x_{i} - \vec{X_{i}}^{k}\right\|_{2}^{2} \right),\forall i\nonumber\\
&\lambda_{i}^{k+1} \leftarrow \lambda_{i}^{k} + \rho \left( x_{i}^{k+1} - \vec{X_{i}}^{k} \right),\forall i \nonumber\\
&X_{l}^{k+1} \leftarrow \frac{1}{N_{l}}\sum_{M(i,j) = l}(x_{i})_{j}^{k+1},\forall l
\end{align}
where superscript $k$ represents the iteration, $N_{l}$ is the number of local copies of the $l$th entry of global consensus variable, $\lambda_{i}$ is the vector of Lagrange multipliers for $i$th subproblem, $M(i,j)$ is the corresponding global consensus entry for $j$th dimension of local variable $x_{i}$, $X_{l}$ denotes the $l$th entry of global consensus variable, and $\rho$ is a step size parameter. The update of $X^{k+1}$ can be viewed as averaging local copies in subproblems. 

\begin{figure}
\centering
\epsfig{file=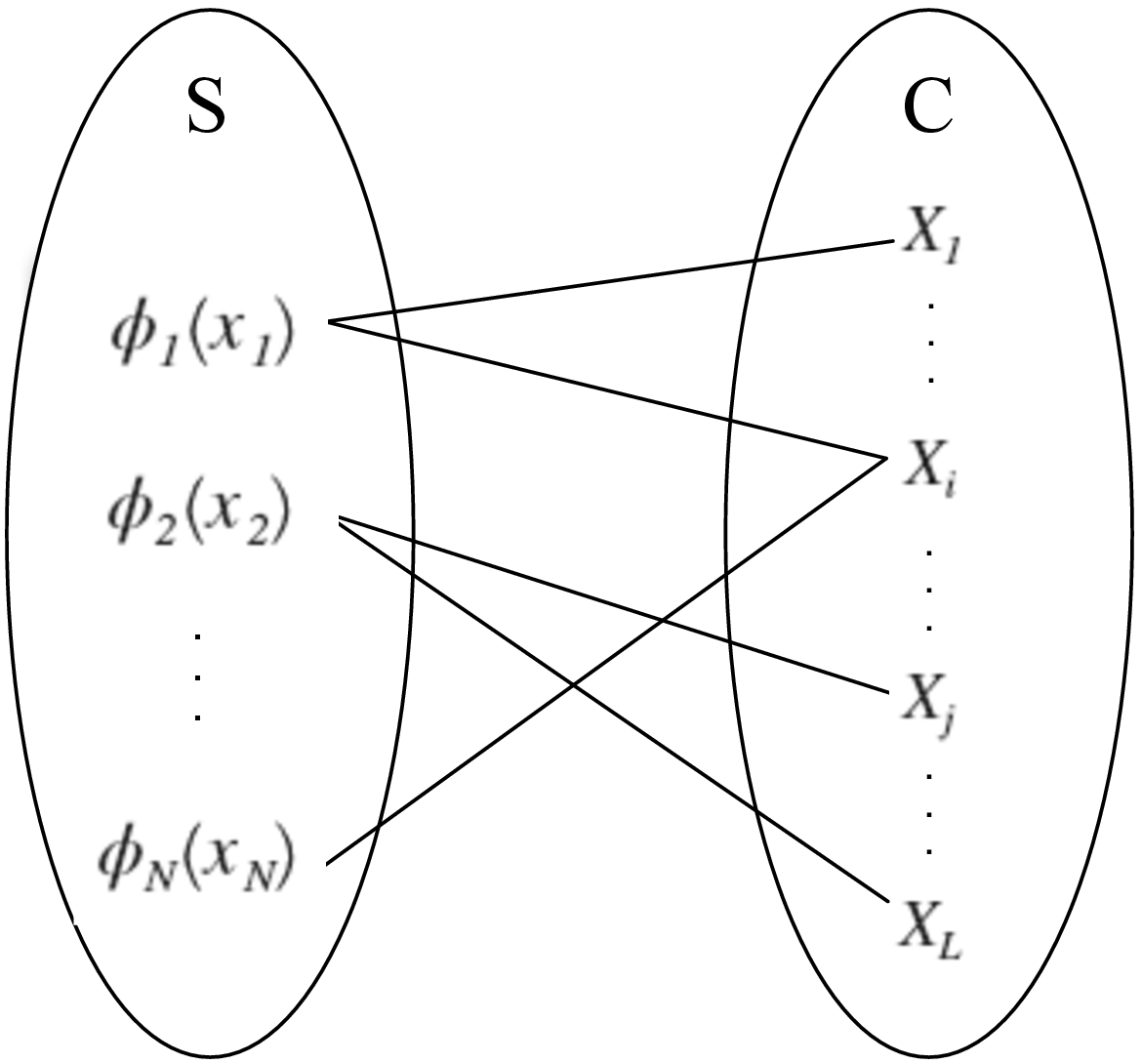, width=1.8in}
\caption{Bipartite graph abstraction for ADMM consensus optimization.}
\label{fig:topology}
\vspace{-10pt}
\end{figure}

Equation \ref{eq:admm} shows that $N$ subproblems can be solved independently. This general form of consensus optimization defines a bipartite graph structure $G(S, C, E)$ where $S$ denotes the set of subproblems containing local variables $\{x_i | i = 1, 2, \ldots, N\}$, $C$ represents the set of consensus variable entries, and $E$ expresses the dependencies as shown in Figure~\ref{fig:topology}. Each subproblem $\phi_{i}(x_{i})$ is connected to its dependent consensus variables, while each consensus variable $X_j$ is connected to subproblems containing its local copies. 

As described above, the computation required for solving a subproblem 
depends on only a subset of the consensus variable
nodes, while the computation needed for a consensus variable requires
information about only a subset of the local variables.  This dependency
relationship makes ADMM computation well-suited for vertex-processing
parallelization. 


\subsection{Vertex Programming Frameworks}

Recent development of vertex programming frameworks, such as Pregel \cite{Czajkowski:Pregel} and GraphLab \cite{YuchengLow:DistributedGraphlab}, are aimed at improving the scalability of graph processing. 
Vertex-centric models in these systems execute user-defined functions on each vertex independently and define the order of execution of vertices. Pregel and GraphLab have superior computational performance over MapReduce for many data mining and machine learning algorithms, such as belief propagation, Gibbs sampling, and PageRank \cite{Czajkowski:Pregel,YuchengLow:DistributedGraphlab,Carlos:GraphlabFramework}. 

In this paper, we implement our synchronous vertex programs in GraphLab.

\subsection{GAS Vertex Programming API}
\label{subsection:api}

Gonzalez et al. \cite{gonzalez2012powergraph} propose the
\emph{gather-apply-scatter} (\emph{GAS}) abstraction that describes
common structures of various vertex programming frameworks.
In the vertex programming
setting, the user defines a data graph with data structures
representing vertices, edges, and messages. The user provides a vertex
program associated with each vertex. The GAS model abstracts 
the program into three conceptual phases of execution
on each vertex. In the \emph{gather} phase, each vertex is
able to aggregate neighborhood information, which could be pushed or
pulled from adjacent nodes. The aggregation
during this phase is user-defined, but must be commutative and
associative. In the \emph{apply} phase, each vertex can use its
aggregated value to update its own associated data. Finally in the
\emph{scatter} phase, each vertex either sends messages to its neighbors or
updates other vertices or edges in its neighborhood 
via global state variables.





\section{Distributed ADMM-based consensus optimization implementation (\ACO)}
\label{sec:algorithm}

In order to implement a graph algorithm using vertex programming, one
needs to define the data graph structure including the vertex, edge
and message data types, as well as a vertex program that defines the
computation.

\subsection{Data Graph and Data Types}
Recall that in ADMM-based consensus optimization, the computation
graph is based on the dependencies between subproblems and consensus
variables described by the bipartite dependency graph $G(S, C, E)$ 
(\figref{fig:topology}). In this graph, any consensus
variable has a degree of at least two, since any local variables that
only appear in one subproblem do not need consensus nodes.

For each node in the bipartite graph, we construct a vertex in the
vertex program.  We use different data types for subproblem vertices
and consensus-variable vertices, denoted $sub$ and $con$,
respectively.  For each subproblem vertex $v_i \in S$, we maintain the
involved local variables in $x_i$ and associated Lagrange multiplier
$\lambda_{i}$, both of which are $n_i$ dimension vectors. Each $v_i
\in S$ also stores a vector $\vec{X_{i}}$ of dimensionality
$|E_{\textbf{v}_{i}}|$ for holding the dependent consensus variable
values. For each consensus variable node $v_j \in C$, we only store
its current value.


\subsection{\ACO\ Vertex Program}
We use the GAS abstraction introduced in Section \ref{subsection:api} 
to describe our ADMM-based consensus optimization implementation, \ACO, shown in Alg.~\ref{alg:main}. 
In each iteration, we define a temporary consensus-variable key-value table \texttt{consensus\_var}, where the key is a consensus variable's global unique id, i.e., \texttt{consensus\_var}$[id(\vec{X})] \rightarrow \vec{X}$. We also define a program variable \texttt{local\_copy\_sum} to aggregate the sum of each local copy of a consensus variable.

\begin{lstlisting}[caption={The \ACO\ vertex program for ADMM-based Consensus \\Optimization on vertex $\textbf{v}_i$ at iteration $k+1$}, label=alg:main, xleftmargin=2.5ex]
%*\ACO\ Algorithm*)
// gather neighbor information
gather(%*$v_{i}$*), %*$(v_{i},v_{j})$*), %*$v_{j}$*)):
	if %*$v_i.type$ == sub*)
		%*\texttt{consensus\_var}$[id(v_j.\vec{X_{j}})]$ $\leftarrow v_j.\vec{X_{j}}^{k}$*)
	else 
		%*\texttt{local\_copy\_sum} $+=$ $v_j.x_j^k[id(v_i.\vec{X_{i}})]$*)
// update the vertex data of v_i
apply(%*$v_{i}$*), %*\texttt{sum\_result}*)):
	// get consensus_var new value, solve objective, update multiplier
	if %*$v_i.type$ == sub*)
		%*$v_i.\vec{X_{i}} \leftarrow \texttt{consensus\_var}$*) 
		%*$v_i.x_{i} \leftarrow \argmin_{x_{i}}(\phi_{i}(x_{i})+ v_i.\lambda_{i}\cdot x_{i} + \frac{\rho}{2}\left\| x_{i} - v_i.\vec{X_{i}}\right\|_{2}^{2})$*)
		%*$v_i.\lambda_{i} \leftarrow v_i.\lambda_{i} + \rho(v_i.x_{i} - v_i.\vec{X_{i}})$ *)
	// average the sum of each local copy
	else 
		%*$v_i.\vec{X_{i}} \leftarrow \frac{\texttt{local\_copy\_sum}}{\mathrm{degree}(v_i)}$*)
// update neighborhood
scatter(%*$v_{i}$*), %*$(v_{i},v_{j})$*), %*$v_{j}$*)):
	// notice consensus node the value change
	if %*$v_i.type$ == sub*)
		notify(%*$v_{j}$*))
	else
		if (convergence_check() == false)
			notify(%*$v_{j}$*)) 

\end{lstlisting}

As shown in Alg.\ref{alg:main}, in the gather, apply and scatter stages, we alternate computation on the subproblem nodes $S$ and the consensus nodes $C$. We describe the computation and communication for each node type below.

\subsubsection{Subproblem Nodes} 
In the gather phase of the $(k+1)^{th}$ iteration, each subproblem node $v_i \in S$ reads the consensus variables updated in $k^{th}$ iteration in its neighborhood. We store each consensus variable in the key-value table \texttt{consensus\_var}. The commutative and associative aggregation function here combines the key-value tables. After getting the updated consensus-variable table, we use it to solve the optimization subproblem in line 13, and the $x_i$ vector is updated to the solution. Note that the subproblem solver is application-specific and is defined by the user. In the scatter phase, the subproblem notifies dependent consensus nodes if $x_i$ was updated. 

\subsubsection{Consensus Variable Node}
The consensus variable node $v_j \in C$ behaves differently in the vertex program. It aggregates all local copies of it from subproblem nodes using summation in gather phase, then update itself with the average value in apply method. In the scatter phase, convergence conditions are used to determine whether related sub problems need to be scheduled to run again. 

\subsubsection{Termination Conditions}

One possible criterion for convergence is the global primal and dual
residual of all consensus variables and their local copies. At the
superstep, an aggregator can be used to aggregate residuals across all
consensus variables. If both primal and dual residuals are small
enough, then we have reached global convergence and the program
stops. If not, all subproblems will be scheduled again in the next
iteration. However, this global convergence criterion
has two disadvantages: first, the use of aggregator will bring
overhead as it needs to aggregate information from all distributed
machines; second, some consensus variables and their
corresponding local copies do not change much and their
subproblem counterparts are still scheduled to run, wasting
computation resources.

Instead, our proposed convergence criterion measures local convergence.
In this local criterion, we check the primal residual and dual residual for
local copies of a consensus variable only. Each consensus vertex
calculates both primal residual and dual residual using its dependent
local copies. If both of them are small, subproblems connecting to
this consensus variable will not be notified in the following 
iteration. 
The notifications to a
particular subproblems come from all consensus variables connecting to
that particular subproblem vertex. If none of the connected consensus
variables notify it, the subproblem skips the following iteration, thus saving computation. A
skipped subproblem node will be notified again if its dependent
consensus variable is updated and the convergence criterion is not
met. 



%
%

\section{Hypergraph partitioning}
\label{sec:partitioning}


In this section, we present our new hypergraph-based partitioning scheme (\hyper) that is better suited for \ACO\ than current state-of-the-art approaches. The primary factors for efficient implementation of distributed graph algorithms are load balancing and communication cost. Vertex programming frameworks ensure good load balancing using a balanced $p$-way cut of the graph. Significant communication costs result from vertices whose neighbors are assigned to different machines. Pregel \cite{Czajkowski:Pregel} uses edge cut, mirrors vertices, and proposes message combiners to reduce communication, while GraphLab uses vertex-cut. An edge-cut has been proven to be convertible to a vertex-cut with less communication overhead \cite{gonzalez2012powergraph}. Therefore, we focus on vertex partitioning in the rest of our discussion in this section. 

\subsection{Problem Definition and Notation} 
Let $G(V,E)$ be a general graph, $\beta$ be a parameter determining the imbalance. For any $v \in V$, let $A(v)$ denote the subset of $M$ machines that vertex $v$ is assigned to. Then the \emph{balanced $p$-way vertex cut problem} for $M$ machines is defined as
\begin{equation}
\begin{aligned}
& \min_{A}
& & \frac{1}{|V|}\sum_{v\in V}|A(v)|\\
& \text{subject to}
& & |\{e\in E| A(e) = m\}| \leq \beta\frac{|E|}{M},\forall m
\end{aligned}
\end{equation}
where $m \in \{1, \ldots, M\}$.
The objective corresponds to the \emph{replication factor}, or how many copies of each node exist across all machines, and the constraint corresponds to a limit on the edges that can be assigned to any one machine.

\subsection{Intuition and State-of-the-Art}
The current state-of-the-art strategy used in GraphLab is a sequential greedy heuristic algorithm \cite{gonzalez2012powergraph}, which we refer to as \greedy. 
Multiple machines process sets of edges one by one and 
place each of them into a machine,    
where the placement $A(v)$ is maintained across multiple machines. When a machine places an edge $(u, v)$, the \greedy\ strategy follows
heuristic rules: if both $A(u)$ and $A(v)$ are $\emptyset$, edge
$(u,v)$ is placed on the machine with the fewest assigned edges; if
only one of $A(u)$ and $A(v)$ is not $\emptyset$, say $A(u)$, then $(u,v)$
is put in one machine in $A(u)$; if $A(u) \cap A(v)\ \neq \emptyset$,
then $(u,v)$ is assigned to one of the machines in the intersection;
the last case is both $A(u)$ and $A(v)$ are not $\emptyset$, but $A(u)
\cap A(v) = \emptyset$, then $(u,v)$ is assigned to one of the
machines from the vertex with the most unassigned edges. 



As we will show, the \greedy\ strategy does not work well with the
\ACO\ bipartite graphs. In \ACO\ and other similar problem structures,
subproblem nodes tend to have much lower degree than consensus
nodes. Because the last heuristic in the greedy scheme is biased to large degree
nodes, \greedy\ places a large number of subproblem nodes onto different machines.

In practice, large-scale \ACO\ involves millions of consensus variables, so computing a high-quality partitioning is more important than fast sequential partitioning. Once partitioned, the same topology may be reused multiple times, for example when performing parameter optimization. In the rest of our discussion, we investigate properties of the \ACO\ bipartite graph and propose a novel and efficient partitioning scheme.

\subsection{Specific Properties of  Consensus Optimization}
We assume the bipartite graph $G(S, C, E)$ of the \ACO\  exhibits four characteristics:
\begin{enumerate}
	\item The consensus variable nodes have a power-law degree distribution $P(d) \propto d^{-\alpha}$, where $\alpha$ is a shape parameter.
	\item The subproblem degree distribution is centered around some small number. We uses a Poisson distribution for the simplicity of analysis, i.e., $P(d) \propto \frac{\lambda^{k}e^{-\lambda}}{d!}$.
	\item For large-scale optimizations, the number of subproblems is larger than the number of consensus variables, i.e., $|S| > |C|$.
	\item The algorithm strictly follows the bipartite structure, i.e., the computation of $S$ only depends on $C$ and the computation of $C$ only depends on $S$.
\end{enumerate}

We briefly justify each of these characteristics. The first characteristic above results from natural power-law degree distributions found in real-data applications.  
Such degree distributions are common in large-scale \ACO\ problems when the variables involved in subproblems correspond to objects in the real world, especially for applications on social and natural networks. 

The second characteristic is a standard requirement for the utility of the ADMM decomposition, where the original optimization problem decomposes into small subproblems that are each easy to solve. Thus the degree of each subproblem node (the number of variables it involves) is small. The Poisson distribution is commonly used to model the number of events occurring during a fixed interval or space. Therefore, it is suitable to describe the number of variables that are involved in a subproblem. The parameter $\lambda$ describes the average number of variables in a subproblem and should be small. 

The third characteristic similarly corresponds to the utility of consensus optimization. The rich models we aim to reason over typically include overlapping interactions among variables. These overlapping interactions make the distribution complex and thus make direct optimization cumbersome. By decomposing the problem into many subproblems that share a smaller set of variables, inference becomes easier. 
In practice, the number of subproblems is roughly an order of magnitude greater than the number of variables. 

The fourth characteristic, that the computation of nodes in $S$ depends only on the values of a subset of $C$ and the computation of nodes in $C$ depends only on the values of a subset of $S$, exhibits itself in many factor-based representations of probabilistic models. This characteristic provides us with extra information about the structure of computation and, when combined with the third characteristic, motivates partitioning only the consensus variables $C$ instead of partitioning over the whole set of nodes $S \cup C$, which inevitably introduces expensive redundancy.

\subsection{Analysis of \ACO }
We first analyze \random\ over all nodes in $G$ and show its inefficiency in 
bipartite graphs with the four characteristics mentioned above, then we present our novel partitioning strategy. If we use random vertex-cut without considering the data dependency structure, the expected number of replications of a random vertex-cut on the bipartite graph is
\begin{align}\label{eq:random_partition_replica}
 \frac{1}{|V|} \mathrm{E} \left[ \sum_{v\in V}|A(v)| \right]  = \frac{M}{|V|} \sum_{v\in C}\left(1 - \mathrm{E} \left[1 - \frac{1}{M}\right]^{d_{1}(v)} \right) \nonumber\\
 + \frac{M}{|V|} \sum_{v\in S} \left(1 - \mathrm{E} \left[1 - \frac{1}{M} \right]^{d_{2}(v)}\right),\nonumber
\end{align}   
where $V = S \cup C$, $A(v)$ is the set of machines on which variable $v$ is located, $M$ is the total number of machines, $d_{1}(v)$ is a consensus variable's degree (power-law-distributed), and $d_{2}(v)$ is a subproblem's degree (Poisson-distributed). This analysis is direct extension of that in \cite{gonzalez2012powergraph}. The factors affecting the expected replication factor are the number of machines, the parameter of power-law distribution $\alpha$, and the parameter of Poisson distribution $\lambda$. We evaluate how these parameters affect the replication factor on synthetic data in our experiments (\secref{sec:experiments}).

If we further assume that each subproblem itself is not very complex and involves just a constant number $c$ of variables, the expected number of replications is: 
\begin{equation}\label{eq:rp random_partition_replica2}
\begin{aligned} 
\frac{1}{|V|} \mathrm{E} \left[\sum_{v\in V}|A(v)| \right]  &= \frac{M}{|V|}\sum_{v\in C}\left(1 - \mathrm{E} \left[1 - \frac{1}{M} \right]^{d_{1}(v)} \right) \\
 & + \frac{M}{|V|}|S|\left(1 - (1 - \frac{1}{M})^c\right)\nonumber
\end{aligned}
\end{equation}
Thus the expected replication factor increases linearly with the number of subproblems $|S|$. 


\subsection{Hypergraph-based Bipartite Graph Partitioning}
Since the random strategy is agnostic to the graph structure, it should be possible to exploit the known bipartite graph structure to improve efficiency. In particular, the bipartite structure suggests that we should aim to partition the consensus variables only. The objective function of the vertex partition over a bipartite graph is defined as
\begin{equation}
\begin{aligned}
& \min_{A}
& & \frac{1}{|V|}\sum_{v\in V}|A(v)|\\
& \text{subject to}
& & |\{e\in E| A(e) = m\}| \leq \beta\frac{|E|}{M},\forall m\\
&&& A(v) = \emptyset,\forall v\in L
\end{aligned}
\end{equation}
When assigning a consensus variable node a machine, we also assign the associated subproblems to that same machine. Because we are cutting only the consensus variable nodes, for each edge located on a machine, the workload added to that machine is the load of the corresponding subproblems plus the workload of consensus variable, which only involves a simple averaging. Therefore, the edge balance in the equation above is equivalent to subproblem node balance, which can be reduced to the following constrained optimization:
\begin{equation}
\begin{aligned}
& \min_{A}
& & \frac{1}{|V|}\sum_{v\in V}|A(v)|\\
& \text{subject to}
& & |\{v\in S| B(v, A) = m\}| \leq \beta\frac{|S|}{M},\forall m\\
&&& A(v) = \emptyset,\forall v\in S
\end{aligned}
\end{equation}
where $A$ is the assignment of consensus variables to machines, $B$ is the mapping from subproblems to machines when the assignment $A$ is given, and $\beta$ is the imbalance factor.

We perform vertex cut only on consensus nodes by treating it as a hyperedge partitioning in the hypergraph view of the bipartite graph. The bipartite graph is converted into a hypergraph $H$ as follows: each node $v\in S$ is a also a node in $H$; each node $v\in C$ is a hyperedge connecting the set of nodes in $H$ corresponding to $v$'s neighbors in $G$. We thus denote the hypergraph as $H = (S, E_{h})$ where $E_h$ is the hyperedge set. Immediately, one can see that the vertex cut of only consensus nodes in $G$ reduces to \hyper\ in $H$. 
    
The hyperedge partitioning problem has been well-studied and there are various packages that can perform hyperedge partitioning efficiently. In our experiments, we use \emph{hMETIS} \cite{Vipin:MultiplevelKwayHypergraphPartitioning}. The objective function is to minimize the sum of external degrees: $\sum_{i=1}^M |E(P_{i})|$, where $P_{i}$ is the set of subproblems that are assigned to the $i$th machine, $|E(P_{i})|$ is the external degree of partition $P_{i}$, i.e., the number of hyperedges that are incident but not fully inside $P_{i}$. Because consensus nodes in $G$ have a one-to-one mapping to hyperedges in $H$, minimizing the sum of external degrees in $H$ is equivalent to minimizing the replication factor in $G$.

\section{Experiments}
\label{sec:experiments}
\begin{figure*}[t!]
\subfigure[Varying $m$ ($\alpha=2$, $\lambda=2$)]{
	\includegraphics[totalheight=0.13\textheight]{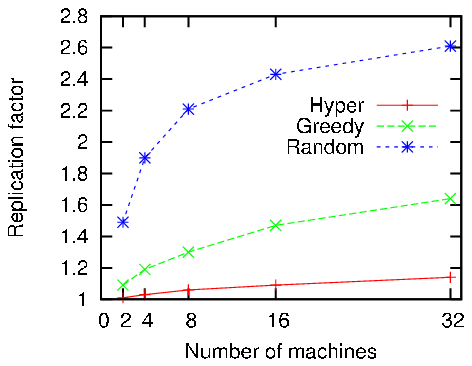} 
	\label{fig:vary_machine_a2l2}
}
\subfigure[Varying $\alpha$ ($\lambda=2$, $m=32$)]{
	\includegraphics[totalheight=0.13 \textheight]{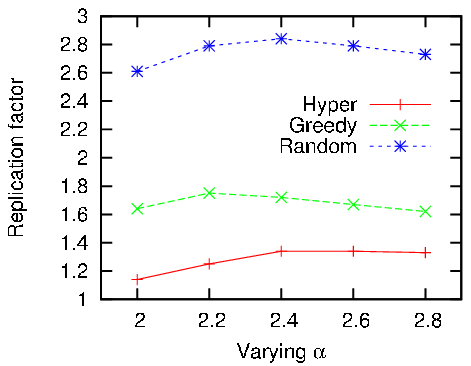}
	\label{fig:vary_alpha_l2m32}
}
\subfigure[Varying $\lambda$ ($\alpha=2$, $m=32$)]{
	\includegraphics[totalheight=0.13\textheight]{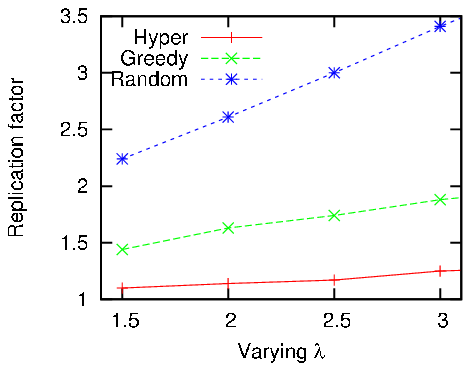}
	\label{fig:vary_lambda_a2m32}
}
\subfigure[Varying $\frac{|C|}{|S|}$ ($m=32$)]{
	\includegraphics[totalheight=0.13\textheight]{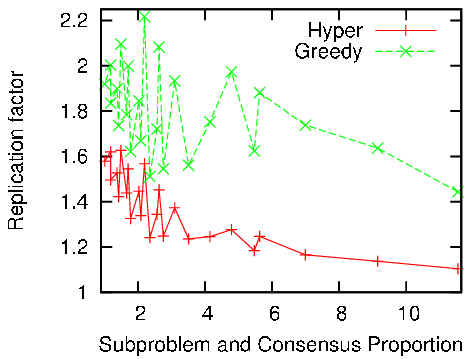}
	\label{fig:analyze_proportion}
}
\label{fig:partitioning_full}
\caption{Comparison between partitioning techniques: \hyper, \greedy, and \sc Rand}
\end{figure*}

In this section, we first compare our hypergraph partitioning vertex-cut technique 
\hyper\
with the greedy vertex-cut algorithm \greedy\
and the hash-based random partitioning \random\
introduced in~\secref{sec:partitioning}.
We then present the evaluation of our \ACO\ implementation using GraphLab on the large-scale 
social network analysis problem introduced in Section~\ref{sec:motivation}. 
In all of our experiments, we use \emph{hMetis} \cite{code:hmetis} with unbalanced factor $\beta=2$ and 
use the \emph{sum of external degree} objective to perform the hyperedge cut. 
Code and data for all experiments will be made available at \url{http://linqs.cs.umd.edu/admm}.

\subsection{Evaluation of Partitioning Strategies}

\setlength{\tabcolsep}{.50em}
\begin{table}[b!]
\centering
\begin{tabular}{cccccccc}
\toprule
\multirow{2}{*}{$\alpha$} & \multirow{2}{*}{$\lambda$} & \multirow{2}{*}{$|S \cup C|$}
  & \multirow{2}{*}{$|E|$} & \multirow{2}{*}{$\frac{|S|}{|C|}$} & \multicolumn{3}{c}{Replication factor} \\ \cline{6-8}
 & & & & & \hyper \hspace{-.1in} & \greedy \hspace{-.1in} & \random \\
\midrule
\multirow{5}{*}{2.0}& 1.5 & 1,254,452 & 1,811,449 & 11.54 & 1.10 & 1.44 &2.24\\
& 2.0 & 1,015,092 & 1,661,788 & 9.15 & 1.14 & 1.64 & 2.61\\
& 2.5 & 799,850 & 1,389,912 & 7.00 & 1.17 & 1.74 & 3.00\\
& 3.0 & 662,938 & 1,247,468 & 5.63 & 1.25 & 1.88 &3.41\\
& 3.5 & 578,983 & 1,142,410 & 4.79 & 1.28 & 1.97 &3.78\\
\midrule
\multirow{5}{*}{2.2} & 1.5 & 647,396 & 1,051,772 & 5.47 & 1.18 & 1.62 & 2.44\\
& 2.0 & 514,906 & 902,526 & 4.15 & 1.25 & 1.75 & 2.79\\
& 2.5 & 409,645 & 792,021 & 3.10 & 1.37 & 1.93 & 3.16\\
& 3.0 & 363,194 & 756,398 & 2.63 & 1.45 & 2.08 & 3.48\\
& 3.5 & 319,539 & 708,340 & 2.20 & 1.57 & 2.22 & 3.80\\
\midrule
\multirow{5}{*}{2.4} & 1.5 & 450,976 & 704,064 & 3.51 & 1.24 & 1.56 & 2.53\\
 & 2.0 & 356,921 & 614,450 & 2.57 & 1.34 & 1.72 & 2.84\\
 & 2.5 & 303,164 & 559,470 & 2.03 & 1.45 & 1.85 & 3.13\\
 & 3.0 & 271,035 & 541,912 & 1.71 & 1.55 & 2.00 & 3.40\\
 & 3.5 & 249,170 & 522,232 & 1.49 & 1.63 & 2.10 & 3.65\\
\midrule
\multirow{5}{*}{2.6} & 1.5 & 375,734 & 580,372 & 2.76 & 1.25 & 1.54 & 2.53\\
 & 2.0 & 308,738 & 515,364 & 2.09 & 1.34 & 1.67 & 2.79\\
 & 2.5 & 265,934 & 474,523 & 1.66 & 1.44 & 1.78 & 3.04\\
 & 3.0 & 237,711 & 451,492 & 1.38 & 1.53 & 1.90 & 3.27\\
 & 3.5 & 218,271 & 437,161 & 1.18 & 1.62 & 2.00 & 3.49\\
\midrule
\multirow{5}{*}{2.8} & 1.5 & 335,411 & 507,033 & 2.35 & 1.24 & 1.51 & 2.50 \\
 & 2.0 & 278,068 & 450,796 & 1.78 & 1.33 & 1.62 & 2.73 \\
 & 2.5 & 242,126 & 420,322 & 1.42 & 1.42 & 1.74 & 2.96 \\
 & 3.0 & 218,350 & 400,794 & 1.18 & 1.50 & 1.84 & 3.14 \\
 & 3.5 & 200,716 & 385,167 & 1.01 & 1.58 & 1.92 & 3.33 \\
\bottomrule
\end{tabular}
\caption{Summary of Synthetic Bipartite Graph with Replication Factor Results ($m=32$)}
\label{tb:dataset}
\vspace{-15pt}
\end{table}


\subsubsection{Dataset Description}

We begin by studying the effect of different partitioning strategies 
on the replication factor of the vertex program.
To produce a dataset that
embodies the rich characteristics discussed in Section
\ref{sec:partitioning}, we generate graphs with a power-law degree
distribution for consensus variable vertices with parameter $\alpha$
and a Poisson distribution over the degree of subproblems with
parameter $\lambda$. We use the random bipartite generator in the
python package \emph{NetworkX} \cite{code:networkx}, which generates
random bipartite graphs from two given degree sequences.

We fix the number of consensus variable vertices to $100,000$ and
generate two degree sequences with varying $\alpha$ and
$\lambda$. Since $\alpha$ in natural graphs is roughly 2.2
\cite{3faloutsos1999power} and subproblems from real problems tend to
be small, we vary $\alpha$ from 2.0 to 2.8, and vary $\lambda$ from
1.5 to 3.5. We use rejection sampling to remove samples with any nodes
of degree less than 2. Because the sum of degrees for $|S|$ and $|C|$
should be equal, their proportion ($|S| / |C|$) can be derived from
$\alpha$, $\lambda$, and $|C|$. We list the properties of generated datasets in Table
\ref{tb:dataset}, where $|S \cup C|$ is the number of vertices and
$|E|$ shows number of edges.

\subsubsection{Replication Factor Results with Synthetic Data}

Given the dataset listed in Table \ref{tb:dataset}, we vary the number of machines 
(partitions) 
$m \in \{2,4,8,16,32\}$ and 
measure the replication factor $RF =
\frac{1}{|V|}\sum_{i=1}^{N}|A(v)|$ of each scheme, denoted as \hyper,
\greedy, and \random\ accordingly in Table~\ref{tb:dataset}. The
parameters' default values are set to: $\alpha=2$, $\lambda=2$, $m =
32$. We list the results in the last three columns of Table
\ref{tb:dataset} and in \figref{fig:partitioning_full}. In general, in all generated
datasets, \hyper\ always has a smaller replication factor than \greedy\ and \random. In the worst
case, \greedy replicates around $17\times$ more vertices than
\hyper\ ($\alpha=2, \lambda=2.5, m=2$), and always replicates
$1.6\times$ more ($\alpha=2.8, \lambda=3.5, m=32$).

\begin{figure*}[t!]
\centering
\subfigure[$SN_{1M}$ (fits on one machine)]{
	\includegraphics[totalheight=0.13\textheight]{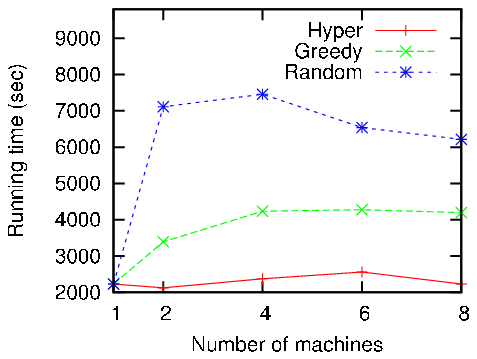} 
	\label{fig:sn_1m_full_conv}
}
\subfigure[$SN_{2M}$]{
	\includegraphics[totalheight=0.13\textheight]{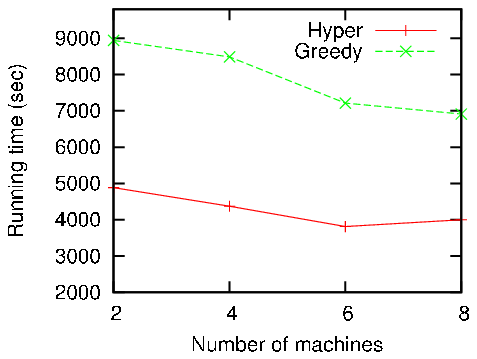}
	\label{fig:sn_2m_full_conv}
}
\subfigure[Weak scaling with increasing size]{
	\includegraphics[totalheight=0.13\textheight]{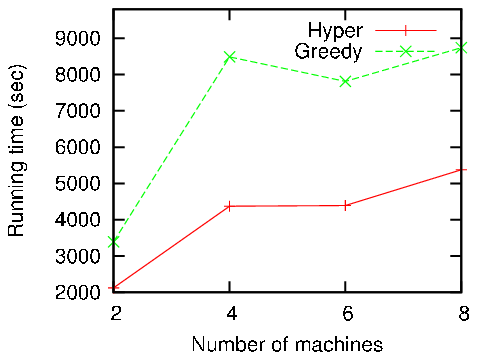}
	\label{fig:weak_scaling}
}
\label{fig:full_conv_time}
\caption{Execution time and weak scaling under full convergence}
\vspace{-15pt}
\end{figure*}


In \figref{fig:vary_machine_a2l2}, we vary $m$ to show how
replication factor grows when the number of machines increases. The results show that \hyper\ is less sensitive to
the number of machines than the other schemes and scales
better in practice.

Next we fix $m = 32$, vary $\alpha$ and $\lambda$ in
\figref{fig:vary_alpha_l2m32} and \figref{fig:vary_lambda_a2m32}
to study the partitioning performance based on different
bipartite graph topologies. Recall that parameter $\alpha$ determines the
power-law shape. As shown in Fig.~\ref{fig:vary_alpha_l2m32}, the
larger $\alpha$ is, the smaller the maximum degree of the
consensus nodes become, and difference between $|S|$ and $|C|$ is smaller,
e.g., when $\alpha = 2.8, \lambda = 2$ in the plot, $|S| / |C|$ is
only $1.78$. In this case, cutting the consensus nodes in
\hyper\ provides less improvement over \greedy.
On the other hand, when $\lambda$ increases, each subproblem has
more variables, and the number of subproblems
decreases. \hyper\ tends to cut more subproblem nodes, as
shown in \figref{fig:vary_lambda_a2m32}.


Finally, in \figref{fig:analyze_proportion}, we plot the relationship
between replication factor and the proportion between $|S|$ and $|C|$. 
When the proportion is small ($< 2$),
\hyper\ and \greedy\ scale similarly, but \hyper\
is still better. As the proportion increases, the advantage of
\hyper\ much more pronounced.

In summary, our proposed hypergraph-based vertex-cut scheme
outperforms the state-of-the-art \greedy scheme provided in GraphLab
implementation \cite{gonzalez2012powergraph,code:graphlabv2} for realistic bipartite graph settings. 
Especially when the two types of nodes in the
bipartite graph are imbalanced, which is typically the case in
consensus optimization, \hyper\ can generate much higher quality
partitions.


\subsection{Performance of \ACO\hspace{0.1mm} for PSL Voter Model}

Next we compare the performance of our proposed \ACO\ vertex-programming
algorithm empirically on an MPI 2 cluster using Open-MPI 1.4.3
consisting of eight Intel Core2 Quad CPU 2.66GHz machines with 4GB RAM 
running Ubuntu 12.04 Linux. We implement our algorithm
using GraphLab 2 (v2.1.4245)
\cite{code:graphlabv2}. 
For each machine in the cluster, we start only
one process with 4 threads (ncpus). We use the synchronous engine
provided by GraphLab 2, which is explained in detail in
\cite{gonzalez2012powergraph}.  
Our proposed approach can be applied
to other vertex programming frameworks easily since it does not use
any special features of GraphLab beyond the synchronous GAS framework.

\subsubsection{Voter Network Dataset Description}

\begin{table}[b]
\vspace{-10pt}
\centering
\begin{tabular}{cccccc}
\toprule
Name & $|S|$ & $|C|$ & $|E|$ & $|S| / |C|$ \\
\midrule
$SN_{1M}$ & 3,307,971  & 1,102,498 & 6,011,257 & 3.00 \\
$SN_{2M}$   & 6,656,775  & 2,101,072 & 12,107,131 & 3.17 \\
$SN_{3M}$ & 9,962,627  & 3,149,103 & 18,113,119 & 3.16 \\
$SN_{4M}$   & 13,349,751 & 4,203,703 & 24,288,223 & 3.18 \\
\bottomrule
\end{tabular}
\caption{Summary of Social Network Data Set for Voter Model}
\label{tb:socialnetdataset}
\end{table}

We generate social voter networks using the synthetic generator of
Broecheler et al. \cite{broecheler:socialcom10,bach:nips12} and create a probabilistic
model using the PSL program in \secref{sec:motivation}. The
details of the datasets are listed in Table
\ref{tb:socialnetdataset}. 
The smallest one $SN_{1M}$ has 6 million
edges and 4.4 million vertices and fits in 4GB memory on a single
machine when loaded in GraphLab; the rest of the datasets do not fit on
a single machine. The
fifth column in Table \ref{tb:socialnetdataset} shows that the
proportion $|S| / |C|$ between two sets in our data graph is around
3. In the voter PSL model, the
variables corresponding to 
the truth of the \textsc{Votes}(person,party) predicate
are consensus variables, and each initialized rule maps to a
subproblem. Each \textsc{Votes}(person, party) appears in
at most eight rules. In practice, PSL programs can be far more
complex and many more subproblems can be grounded, thus the
proportion may be even larger. In such cases, \ACOH\ partitioning will even further reduce communication cost.

\subsubsection{Performance Results with PSL Inference}

\begin{figure}[H]
\vspace{-5pt}
\centering
\epsfig{file=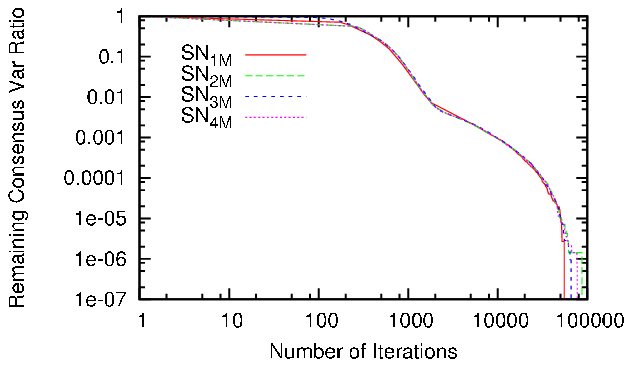, width=2.6in}
\caption{Convergence Rate in PSL Voter Model}
\label{fig:convergence}
\vspace{-10pt}
\end{figure}

To evaluate the performance of our algorithm, we use a GraphLab vertex
program that implements our \ACO\ algorithm described in Section
\ref{sec:algorithm} and vary the partitioning technique. We use the
method described in \cite{bach:nips12} to solve the quadratic
subproblems defined by voter PSL program.  We consider performance of
\ACO\ under two settings: full convergence and early stopping when one
considers computation time budgets. It is important to consider the early-stopped
setting since \ACO\ is known to have very fast initial convergence and
then slow convergence toward the final optimum
\cite{Boyd:DistrubutedADMM}. In practice, one can stop early when the
majority of variables have converged and quickly obtain a high-quality
approximate solution. As shown in \figref{fig:convergence},
inference in the PSL voter model quickly converges on 99\% of the
consensus variables, taking 1,000 iterations on all four datasets.

\textbf{Full convergence:} 
As shown in \figref{fig:sn_1m_full_conv} and
\ref{fig:sn_2m_full_conv}, we first vary $m$ to show the running time
and speedup under the full convergence setting. Because $SN_{1M}$ is
able to fit into memory on one machine, communication cost overwhelms
extra computation resources, and prevents distributed computation from
performing better than single machine. In
Fig.~\ref{fig:sn_1m_full_conv}, \ACOG\ and \ACOR\ perform 
($2\times$ to $4\times$) worse than the single machine setting. \ACOH\ has
similar running times to those of a single machine and is $2\times$
better than \ACOG.

On larger data sets that cannot fit on a single machine, our approach
is approximately twice as fast as
\ACOG\ (Fig.~\ref{fig:sn_2m_full_conv} and
\ref{fig:weak_scaling}). However, the speedup for the full convergence
setting is not as significant because some consensus variables take many
iterations to converge. Fewer than 1\% of the consensus variables
are still active after 1,000 iterations, but as long as any one variable
has not converged, increasing the number of machines will not
produce speedup in terms of computation time.
In \figref{fig:weak_scaling}, we evaluate weak scaling by increasing machine and dataset together.  
Both \hyper\ and \greedy\ scale well on large datasets that cannot fit into one machine.

\eat{
\textbf{hui: working now on the plots and use the following sketches}
\begin{enumerate}
	\item show the run time of 500k, and 1M data, point out two messages: 1) if data can fit in one machine, do not use distributed graphlab, 2) the speed up in general admm consensus optimization will not be good in distributed setting. (point out that ADMM as noted by the authors exhibits very slow convergence (Boyd et al., \cite{Boyd:DistrubutedADMM}) (pgs 6-7) to accurate solutions.) [2 plots] [done, see Fig. 3]
	\item show the weak scale up curve, and mention that we're close to optimal. [1 plot] [done, see Fig. 3] [done]
	\item next motivated by the resource consumption, and slow convergence, show (one iteration speedup), convergence rate, to justify that our method is good. [2 plots] 
\end{enumerate}
}

\textbf{Performance under early stopping:} 
As shown in \figref{fig:convergence}, the majority of nodes converge
quickly. Since modern computing models often include a pay-as-you-go
cost, one may not benefit from waiting for the last few variables to
converge. For instance, the last 1\% of vertices in $SN_{2M}$ take
$2/3$ of the total time for full convergence. Motivated by this
reasoning, in the following experiments, we measure the running time
to complete 1,000 iterations of \ACOH\ and \ACOG, regardless of the
convergence status at the end of the last iteration.

\begin{figure}[b]
\vspace{-15pt}
\centering
\subfigure[Run time of iterations ($SN_{4M}$)]{
	\includegraphics[totalheight=0.124\textheight]{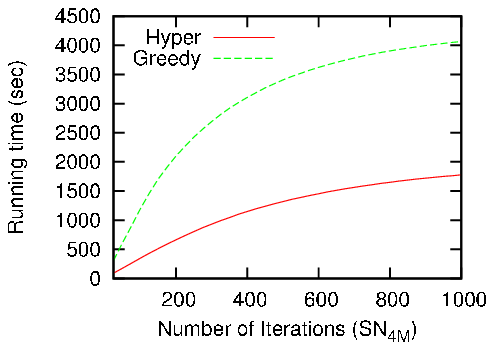}
	\label{fig:1k_scaleup}
}
\subfigure[Execution Time ($SN_{2M}$)]{
	\includegraphics[totalheight=0.124\textheight]{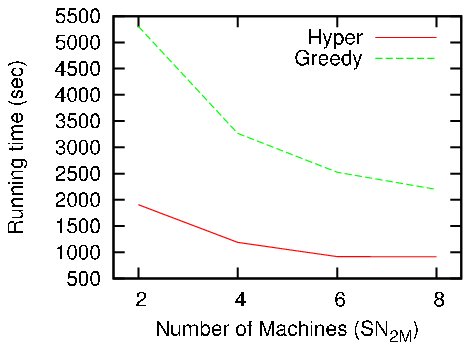} 
	\label{fig:1k_time}
}
\label{fig:1k_evaluation}
\caption{Performance of the first 1000 iterations.}
\end{figure}

In \figref{fig:1k_scaleup}, we show the accumulated running time of
each iteration. Note because we use synchronous setting, both
algorithms have the same state at the end of each iteration. In
\figref{fig:1k_scaleup}, we show that \ACOH\ performs $2\times$ to
$4\times$ better than the \ACOG\ because of the reduction in
communication cost. In Fig.~\ref{fig:1k_time}, we vary the number of
machines for the same dataset $SN_{2M}$ to show the speed up. There
are diminishing returns on increasing the number of machines, due to
the communication overhead incurred by adding machines, but our
hypergraph partitioning produces overall faster computation.

\commentout{
Since the computation of \ACO\ exhibits bipartite structure, we test
on randomly generated bipartite graphs. To produce a dataset that
embodies the rich characteristics discussed in Section
\ref{sec:partitioning}, we generate graphs with a power-law degree
distribution for consensus variable vertices with parameter $\alpha$
and a Poisson distribution over the degree of subproblems with
parameter $\lambda$. We use the random bipartite generator in the
python package \emph{NetworkX} \cite{code:networkx}, which generates
random bipartite graphs from two given degree sequences.

We fix the number of consensus variable vertices to $100,000$ and generate two degree sequences with varying $\alpha$ and $\lambda$. Since $\alpha$ in natural graphs is roughly 2.2 \cite{3faloutsos1999power} and subproblems from real problems tend to be small, we vary $\alpha$ from 2.0 to 2.8, and vary $\lambda$ from 1.5 to 3.5. We use rejection sampling to remove samples with any nodes of degree less than 2. Because the sum of degrees for $|S|$ and $|C|$ should be equal, their proportion ($|S| / |C|$) can be derived from $\alpha$, $\lambda$ and $|C|$. We list generated datasets in Table \ref{tb:dataset}, where $|S \cup C|$ is the number of vertices and $|E|$ shows number of edges.

\subsubsection{Replication Factor Results with Synthetic Data}
Given the dataset listed in Table \ref{tb:dataset}, we vary the number of machines (partitions) $m \in [2,4,8,16,32]$ and perform three different partitioning techniques, \ACOH, \ACOG, and \ACOR, and measure the replication factor $RF = \frac{1}{|V|}\sum_{i=1}^{N}|A(v)|$ of each scheme, denoted as \hyper, \greedy\ and \textsc{Rand} accordingly in Table~\ref{tb:dataset}. The parameters' default values are set to: $\alpha=2$, $\lambda=2$, $m = 32$. We list the result in the last three columns of Table \ref{tb:dataset} ($m=32$) and Fig.~3. In general, in all generated datasets, \ACOH\ is always better than \ACOG\ and \ACOR. In the worst case, \greedy\ replicates around $17\times$ more vertices than \hyper\ ($\alpha=2, \lambda=2.5, m=2$), and at least replicates $1.6\times$ more ($\alpha=2.8, \lambda=3.5, m=32$). 

\begin{figure*}[t!]
\centering
\subfigure[$SN_{1m}$ (fits on one machine)]{
	\includegraphics[totalheight=0.13\textheight]{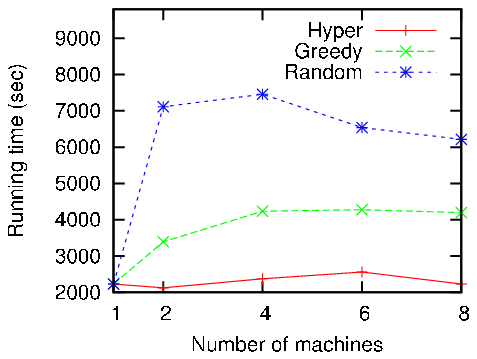} 
	\label{fig:sn_1m_full_conv}
}
\subfigure[$SN_{2m}$]{
	\includegraphics[totalheight=0.13\textheight]{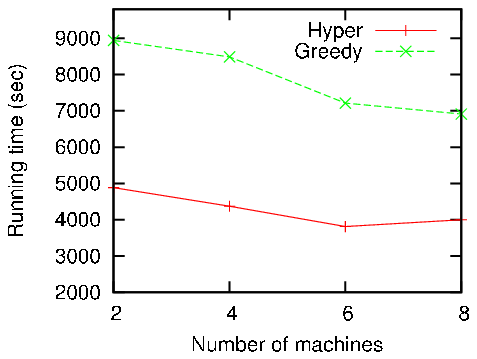}
	\label{fig:sn_2m_full_conv}
}
\subfigure[Weak scaling with increasing size]{
	\includegraphics[totalheight=0.13\textheight]{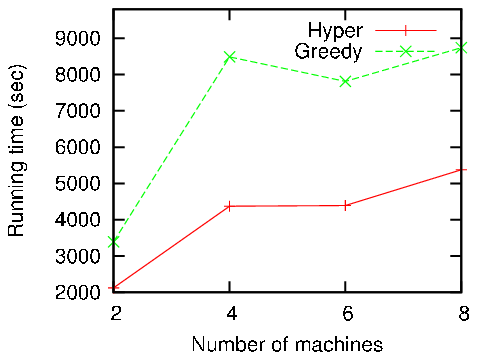}
	\label{fig:weak_scaling}
}
\label{fig:full_conv_time}
\caption{Execution time and weak scaling under full convergence}
\vspace{-15pt}
\end{figure*}


In Fig.~\ref{fig:vary_machine_a2l2}, we vary $m$ to show how replication factor grows when the number of machines increases. From the results, we can conclude that \hyper\ is less sensitive to the number of machines than the other two schemes and has better scale up in practice.

Next we fix $m = 32$, vary $\alpha$ and $\lambda$ in Fig.~\ref{fig:vary_alpha_l2m32} and \ref{fig:vary_lambda_a2m32} accordingly to study the partitioning performance on different bipartite graph topologies. Parameter $\alpha$ determines the power-law shape. As shown in Fig.~\ref{fig:vary_alpha_l2m32}, the larger $\alpha$ is, the smaller the maximum degree becomes of the consensus nodes, and difference between $|S|$ and $|C|$ is smaller, e.g., when $\alpha = 2.8, \lambda = 2$ in the plot, $|S| / |C|$ is only $1.78$. In this case, cutting the consensus nodes in \hyper\ provides less improvement over \greedy\ than  when $\alpha$ is small.  On the other hand, when $\lambda$ increases, sub problem has more variables, and the number of sub problems decreases. \textsc{hyper} tends to cut more sub problem nodes, as shown in Fig.~\ref{fig:vary_lambda_a2m32}. 


Finally, in Fig.~\ref{fig:analyze_proportion}, we plot the relationship between replication factor and the proportion between $|S|$ and $|C|$ using all datasets. When the proportion is small ($< 2$), \textsc{hyper} and \textsc{greedy} scale similarly, but \textsc{hyper} is still better. As the proportion increases, the advantage of \textsc{hyper} becomes much more apparent.

In conclusion, the vertex-cut scheme using our hypergraph formulation performs better than the state-of-the-art scheme, provided in GraphLab implementation \cite{gonzalez2012powergraph, code:graphlabv2}. Especially when the two types of nodes in the bipartite graph are imbalanced, which is typically the case in consensus optimization, \ACOH\ can generate much higher quality partitionions.
}

\section{Related Work}

The ADMM algorithm, recently popularized by Boyd et
al. \cite{Boyd:DistrubutedADMM}, has been used for many applications
such as distributed signal processing \cite{Ernie:ADMM-Bregman}
and inference in graphical models \cite{Tziritas:MRFDD}. In particular,
many large-scale distributed optimization problems can be cast as
consensus optimization and use ADMM to solve them
\cite{Georgios:ADMM-SVM,bach:nips12,bach:uai13}.

Tziratas et al. \cite{Tziritas:MRFDD} proposed a general
messaging-passing scheme for MRF optimization based on dual
decomposition. Their solution has a master-slave MRF structure that is
analogous to our bipartite topology. Building on this work, dual
decomposition has similarly been proposed to perform distributed
inference in Markov Logic Networks (MLNs)
\cite{Shavlik:ScalingInferenceForDD}. This work showed that combining
MRF-level partitioning and program-level partitioning produces
superior performance compared with just MRF-level
partitioning. MRF-level partitioning treats the grounded MLN as a
collection of trees, which is different from our partitioning
objective.

There are several contributions which discuss implementations of ADMM-related
distributed algorithms.  Boyd et
al. \cite{Boyd:DistrubutedADMM} discusses the implementation of ADMM
in MapReduce with global consensus variables, which is a special
case of our setup. GraphLab \cite{graphlab:Code} contains
implementations of the MRF dual decomposition from
\cite{Tziritas:MRFDD}.  Also related, a large-scale implementation of the
``accuracy of the top'' algorithm \cite{Boyd:AccuraryAtTop} 
proposes methods for speeding up convergence
of the top quartile, including tuning the communication topology,
for an ADMM consensus optimization algorithm written in Pregel. 


\section{Conclusion}
\label{sec:conclusion}

In this paper, we introduce a vertex programming algorithm for
distributed ADMM-based consensus optimization. To mitigate
the communication overhead of distributed computation, we provide a novel
partitioning strategy that converts the bipartite computation graph
into a hypergraph and uses a well-studied hypergraph cut algorithm to assign
nodes to machines. This combination of the ADMM vertex program and
hypergraph partitioning enables 
distributed optimization over large-scale data.
Our experiments
on probabilistic inference over large-scale, synthetic social networks
demonstrate that our contributions lead to a significant improvement
in scalability. Additionally, the partitioning scheme is of
independent interest to researchers and practitioners, since many
other graph algorithms also have a bipartite computation structure and
will similarly benefit from the reduced communication overhead induced
by hypergraph partitioning.

\bibliographystyle{IEEEtran}
\def\IEEEbibitemsep{1pt}
\bibliography{IEEEabrv,admm}
%



\end{document}